\documentclass[11pt,a4paper]{article}
\usepackage[hyperref]{naaclhlt2018}
\usepackage{times}
\usepackage{latexsym}
\usepackage{url}

\usepackage[pdftex]{graphicx}
\usepackage{amsmath}
\aclfinalcopy 
\def\aclpaperid{5} 

\title{Improving Retrieval Modeling Using Cross Convolution Networks And Multi Frequency Word Embedding}

\author{Guozhen An$^{1,2,}$\thanks{\; The first two authors contributed equally} $^{\,,\dagger}$, Mehrnoosh Shafiee$^{3,*,}\thanks{\; Work performed as an intern in OATH}$ , Davood Shamsi$^{4}$ \\
$^1$Department of Computer Science, CUNY Graduate Center, USA \\
$^2$Department of Mathematics and Computer Science, York College (CUNY), USA \\
$^3$Department of Electrical Engineering, Columbia University, USA \\
$^4$OATH Inc, USA \\
  {\tt gan@gradcenter.cuny.edu, s.mehrnoosh@columbia.edu}\\         
  {\tt davood.shamsi@oath.com} \\}

\date{}

\begin{document}
\maketitle
\begin{abstract}

To build a satisfying chatbot that has the ability of managing a goal-oriented multi-turn dialogue, accurate modeling of human conversation is crucial. In this paper we concentrate on the task of response selection for multi-turn human-computer conversation with a given context. Previous approaches show weakness in capturing information of rare keywords that appear in either or both context and correct response, and struggle with long input sequences. We propose Cross Convolution Network (CCN) and Multi Frequency word embedding to address both problems. We train several models using the Ubuntu Dialogue dataset which is the largest freely available multi-turn based dialogue corpus. We further build an ensemble model by averaging predictions of multiple models. We achieve a new state-of-the-art on this dataset with considerable improvements compared to previous best results.
  
\end{abstract}

\section{Introduction}
\label{introduction}

One of the primary objectives in Artificial Intelligence (AI) is the task of building a conversational agent that can naturally and coherently communicate with humans. The solution can significantly change the interaction between clients and customers, and has appealing applications to service providers in many different areas. This task can be simplified to define different problems whose solutions move us toward accurate understanding and modeling of human conversation. The two mainstream models can be distinguished as follows: a \textit{Generative model} that tries to generate responses in multi-turn conversations \cite{sordoni2015neural,wen2015semantically,wen2015stochastic,shang2015neural}, and a \textit{Retrieval model} which retrieves potential responses from the massive repository and selects the best one as the output \cite{yan2016learning,ji2014information}. While the first model is a more flexible and powerful model, it is considerably harder to implement. According to the current state of AI, we are far away from a generative model for long and multi-domain conversations.

Until recently, proposed solutions for building dialogue systems required significant hand-engineering of features. This limits the number of responses and situations in which the system can be deployed. More recently, researchers tempt to apply machine and deep learning methods to create a model that can learn the essential information in conversational data. One vital aspect of human conversation is the contextual and semantic relevance among sentences. The sequence modeling approaches are shown to be effective to capture these information. More specifically, Recurrent Neural Networks (RNN) built by Long Short-Term Memory (LSTM) units~\cite{hochreiter1997long} have been effectively utilized for extracting contextual and semantic information in other language related problems such as speech recognition, state tracking, image captioning, etc.~\cite{sutskever2014sequence,cho2014learning,henderson2013deep,graves2013speech,xian2017self}. 

In this paper, we consider the problem of next response ranking for multi-turn human-computer conversation with a given context. The model provides candidates' ranking and selects the one with the highest rank as the next utterance. This problem is an important and challenging task for the retrieval-based dialogue model.

Previous RNN based approaches to response selection take the context and the response candidate as two separate word sequences and feed them to the RNN in order to obtain two embedding vectors. The response is then selected based on the similarity of the candidate embedding with the context embedding \cite{lowe2015ubuntu,kadlec2015improved,baudivs2016sentence,xu2016incorporating}. There are mainly two shortcomings of previous solutions. The first shortcoming corresponds to the method of representing words in the context and candidates. More specifically, in order to efficiently represent words of either the context or the response and feed them to the RNN, we use word vector embeddings. A word embedding is a vector which represents word's semantic and synthetic features. Intuitively, we map words to $N$-dimensional vectors, where vectors that are relatively close represent words with similar or related meaning \cite{mikolov2013distributed}. However, in order to have sufficient semantic and syntactic information for a word, and also to make this method computationally efficient, we require the word to appear in the corpus for at least a certain number of times. Hence, rare words, which are usually technical words and carry important information, are missed by such word embedding method. The second shortcoming relates to the performance of RNN. In fact, LSTM units are vulnerable to losing information when the input sequence is long (which is the case in multi-turn response selection. See Section~\ref{relatedwork}). Furthermore, in the RNN based models, the inputs to the RNN are the sequence of word embeddings of the entire context (or response), and the output is a single vector which represent the contextual and semantic dependency and relevance of the words in the entire context (or response) and does not carry word level information. To address the first problem, we utilize two word embedding layers: one for frequent words and one for rare words. To address the second issue, we extract the similarity between individual words in the context and the response. We note that when an utterance shares a rare word with some context, it is more probable that it is the correct response for the context. Therefore, we design a layer to extract the information of rareness of shared words in the context and the response. 

In this paper, we propose a model that integrates sequence and word level information. We train our model on the Ubuntu Dialogue dataset, which consists of roughly one million two-way conversations extracted from the Ubuntu chat logs. Moreover, this data set is considered to be unstructured dialogues where there is no a priori logical representation for the information exchanged during the conversation \cite{lowe2017training} which is a desirable property to test a retrieval model.

We summarize our contributions as follows:
\begin{itemize}
\item We design Cross Convolution Network (CCN) that gets two inputs (matrix representations of two sentences, feature matrices of two images, etc.) and extracts similarity of the inputs.
\item We propose Multi Frequency Word Embedding that efficiently capture both frequent and rare words of the corpus.
\item Our experimental results show a considerable improvement over previous results on Ubuntu Dialogue Corpus \cite{lowe2015ubuntu}.
\end{itemize}

The remainder of this paper is structured as follows. In Section~\ref{relatedwork}, we review previous work. Description of the dataset can be found in Section~\ref{dataset}. In Section~\ref{method}, we present our methods to capture sequence level and word level information in detail. Section~\ref{experiment} focuses on the experimental setup while our results are presented in Section~\ref{result}. Finally, We conclude and discuss future research directions in Section~\ref{conclusionfuture}. 

\section{Related Work}
\label{relatedwork}

The problem of next response selection in multi-turn conversation is more general than a traditional question answering (QA) problem \cite{yih2015semantic,yu2014deep}. The prediction is made based on the entire conversation context which does not necessarily include a question. In single turn response selection, the model ignores the entire context and only leverages the last utterance to select response \cite{lu2013deep,ji2014information,wang2015syntax}. Since an utterance can change the topic or negate/affirm the previous utterances, it is of paramount importance that models for response selection in multi-turn conversation have a certain understanding of the entire context. Moreover, next response selection system is a supervised dialogue system since it incorporates explicit signals specifying whether the provided response is correct or not \cite{lowe2016evaluation}. This system is of interest because it admits a natural evaluation metric, namely the recall and precision measures (See Section~\ref{dataset} for a detailed explanation.). We consider Ubuntu Dialogue Corpus \cite{lowe2015ubuntu} to evaluate our retrieval-based model since the dataset is the most relevant public dataset to supervised dialogue systems \cite{lowe2016evaluation}.

The original paper that introduced the Ubuntu Dialogue dataset have
implemented a TF-IDF model in addition to neural network models with vanilla RNN and LSTM \cite{lowe2015ubuntu}. Later, \citet{kadlec2015improved} evaluated the performances of various LSTMs, Bi-LSTMs and CNNs (Convolutional Neural Networks \cite{kalchbrenner2014convolutional}) on the dataset and created an ensemble by averaging predictions of multiple models. An RNN-CNN model combined with attention vectors is implemented by \citet{baudivs2016sentence}. Further, Multi-view Response Selection \cite{zhou2016multi} proposed an RNN-CNN model which integrates information from both word sequence view and utterance sequence view. A deep learning model incorporating background knowledge to enhance the sequence semantic modeling ability of LSTM is implemented in \citet{xu2016incorporating} that achieved the state-of-the-art result.

In spite of these efforts, the study of \citet{lowe2016evaluation} found that the automated dialogue systems built using machine and deep learning methods perform worse than human experts in Ubuntu system. This confirms that further investigation in retrieval dialogue system using this dataset is worthwhile and motivates us to conduct this research.

\section{Data}
\label{dataset}

The Ubuntu Dialogue Corpus~\cite{lowe2015ubuntu} is the largest freely available multi-turn based dialogue corpus which consists of almost one million two-way conversations extracted from the Ubuntu chat logs. We use the second version of this dataset in this paper. The dataset was preprocessed as follows: 
\begin{itemize}
\item Named entities were replaced with corresponding tags (name, location, organization, url, path). 
\item Two special symbols; namely, $eou$ and $eot$ are used to denote the end of utterances and turns, respectively. 
\item The training set consists of tuples of the form $<$$context, response, label$$>$, where the $label$ indicates whether the provided $response$ is the correct response for the $context$ or not. For any instance of the form $<$$context, response, 1$$>$, the set includes an instance of the form $<$$context, response^\prime, 0$$>$ where the $response^\prime$ is a randomly sampled utterance from the entire data to create balanced dataset. 
\item The test set is created by $2\%$ of the whole dataset which is approximately 20k instances. Each test instance consists of a $context$ followed by 10 candidate responses with the first candidate being the correct one. The other $9$ responses are drawn randomly from the entire corpus~\cite{lowe2015ubuntu}. Furthermore, a validation set of the same size and structure is provided.
\end{itemize}

The system is required to rank the candidate responses and output the highest ranking. We note that some of the sampled candidates which are labeled as incorrect can be relevant to the context, and hence, considered as correct. Hence, we may examine the system's ranking as correct if the correct response is among the first $k$ candidates. This quantity is denoted by $Recall@k$. Most of the previous papers have used the pairs of $(m,k)$ to be $(10,1)$, $(10,2)$, and $(10,5)$ to report their models' performance~\cite{lowe2015ubuntu,zhou2016multi,kadlec2015improved}.

\section{Method}
\label{method}

In this section, we provide details on the networks and layers we used to build our models. We start by introducing the Cross Convolution Network that captures semantic similarity of the context and response. We then elaborate on Multi Frequency Word Embedding. It is followed by explanation of LSTM network and Common Words Frequency layer.

\subsection{Cross Convolution Network}
\label{cuwordsrelation}
In many instances, a handful number of words reveal the purpose of conversation; therefore, one may expect to see the exact same words in the context or their derivations in the correct response. Our experiments show that RNN models fall short of capturing all these information especially when the input sequence is long. Motivated by this, we design a Cross Convolution Network that intrinsically can be deployed in any classifying problem of a pair of objects. 
We would like to mention that CCN is different with the architecture proposed in \citet{wan2016deep} which utilizes Bi-LSTM and requires to learn parameters of an interaction tensor to capture semantic matching of two sentences.
\begin{figure}[h]
	\centering
    \includegraphics[width=\linewidth]{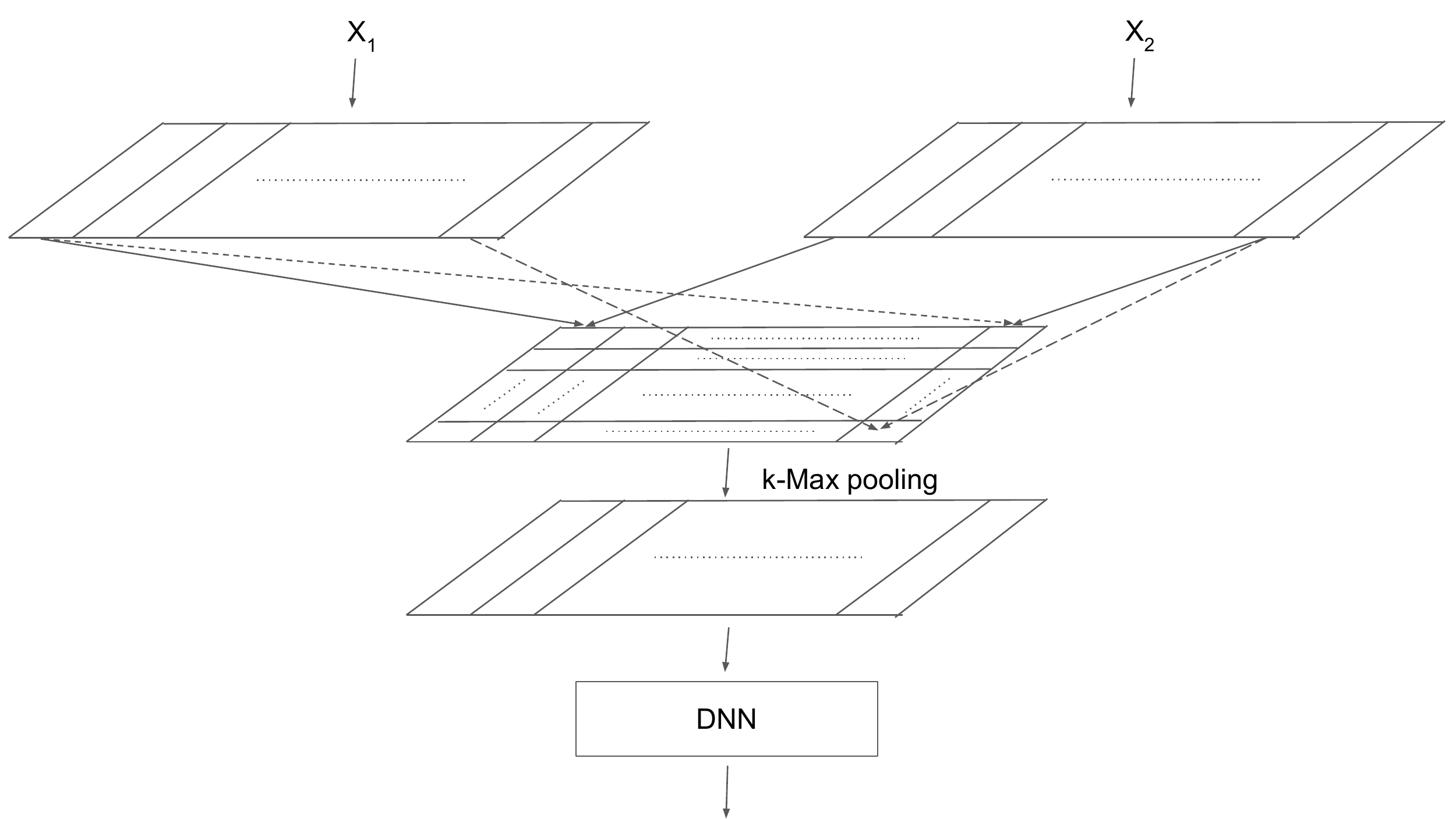}
    \caption{{\it Cross Convolution Network (CCN). $X_1$ and $X_2$ are the inputs to the network. CCN computes convolution of $X_1$ over $X_2$ and passes the output result to a $k$-Max Pooling to extract the first k largest output of each column. A dense layer is then applied to calculate a number that measures similarity of $X_1$ and $X_2$.}}
    \label{fig: CCN}
\end{figure}


At a high level, a Cross Convolution Network accepts two matrices, $X_1$ and $X_2$, and computes convolution of $X_2$ over $X_1$. $k$-Max Pooling is then applied to the output matrix in order to take the $k$ largest element of each of its column. The output is then fed to a Dense layer (vanilla RNN) that measures similarity of $X_1$ and $X_2$. As in Convolutional Neural Network, we need to specify the window and stride sizes for computing the convolution of inputs. Figure~\ref{fig: CCN} shows structure of Cross Convolution Network.

For the task of response selection, we include the following layers to extract the word level information in the context and in the corresponding response:

\textbf{Dot Product Layer} Given the sequence of embedded word vectors of a context and a response, for each word $w_r$ in the response, we calculate its inner product with every word $w_c$ in the context. In other words, we calculate convolution of the context with each of the response words (as Convolutional filter) while window and stride sizes are equal to one.

\textbf{$\bf k$-Max Pooling and Dense Layer} Given the output of the Dot Product Layer, we pick the first $k$ maximum values for each filter $w_r$. We then use a dense layer (DNN with some activation function) to calculate the probability of the corresponding label of the instance to be one.

In matrix representation of the context and response, in which the $i$-th column of the context (response) matrix is the embedding vector representing the $i$-th word in the context (response), we formulate the layer operation as

\begin{equation}
\begin{aligned}
S&=[S_1, S_2, \dots, S_L]=R^T C,\\
\bar{S}_i&=kmax(S_i),\\
s_3&=a^T \times [\bar{S}_1; \bar{S}_2; \dots; \bar{S}_L]+b,\\
p_3&=f(s_3),
\end{aligned}
\end{equation}

where $R$ is the response matrix, $C$ is the context matrix, and $S$ is the dot product output. $kmax (\cdot)$ is $k$-Max Pooling function which picks the first $k$ maximum values of each of the column of matrix $S$. Moreover, $a$ and $b$ are trainable weight vector and bias of the dense layer, respectively. $f(\cdot)$ can be any activation function. $k$ is a hyper-parameter for the model, and $L$ is the maximum number of words in the contexts and responses (the smaller contexts and responses are padded using zero vectors.). 

\subsection{Multi Frequency Word Embedding}
\label{wordembedding}

To have high quality representation of words that capture syntactic and semantic word relationships, we use two types of word embedding layers in our models. As noted in~\citet{lowe2017training}, failure of understanding semantic similarity of context and response is the largest source of error from Dual-LSTM model (see Section~\ref{lstm}).  We observed that our dual LSTM model performed worse when rare words appeared in either or both context and response. One potential explanation is that when training the word embeddings, rare words are removed for the purpose of computational efficiency. However, this weakens the word embeddings due to the loss of information that occurs by ignoring rare words. In order to capture these rare word relation, we utilize multiple word embedding layers instead of one, we attempt to use two word embedding layers, which we refer to as low frequency, and high frequency layers.

Given the word sequences of context and response, words are mapped into $N$-dimensional embedding vectors. While $N$ is a hyper-parameter and needs to be specified, the word embeddings can be initialized with random vectors or with pre-trained word vectors. We use two independent word embedding layers inside a single model. First, we count total appearance of each word in train set of context and response, then filter frequent words and rare words during training from each context and response. 
The high frequency word embedding layer is the same as the word embedding of~\citet{lowe2015ubuntu}, which captures word relation of frequent words and feed to LSTM in future stage to get internal representation of context and response. The low frequency word embedding layer is trained using only rare words from train set of context and response.
We denote the high frequency word filter, low frequency word filter, and embedding layer by $HFWF(\cdot)$, $LFWF(\cdot)$ and $WE(\cdot)$, respectively. Therefore,

\begin{equation}
\begin{aligned}
& WE_h = WE(HFWF(contexts \cup responses))\\
& WE_l = WE(LFWF(contexts \cup responses)),\\
\end{aligned}
\end{equation}

where $WE_h$ ($WE_l$) is the corresponding word embeddings for high (low) frequency word embedding layer that is train on high (low) frequency words in the entire training set.


\subsection{LSTM}
\label{lstm}
\subsubsection{Context and Response Embedding} 

Long Short-Term Memory (LSTM) is well known for capturing information of long sequences. Inspired by~\citet{lowe2015ubuntu}, we use two LSTM networks with shared weights to produce the final representation of context and response by feeding word embeddings one at a time to the respective LSTM. Word embeddings are initialized using the pre-trained $300$-dimensional GloVe Word Vector Model~\cite{pennington2014glove}, and updated during training phase. We use one hidden layer for each LSTM with output size $256$. We denote the LSTM layer by $LSTM(.)$. Therefore,

\begin{equation}
\begin{aligned}
& c = LSTM(WE(context))\\
& r = LSTM(WE(response))\\
\end{aligned}
\end{equation}

where $c$ and $r$ are the final hidden state of LSTM layer. We refer to this model as \emph{Dual-LSTM}. This is the baseline model proposed in~\citet{lowe2015ubuntu} where the response is then selected based on the similarity of $c$ and $r$ which is measured by the inner product of the two embeddings.

Finally, we dense the hidden state of LSTM layer, and calculate the probability of the response to be the correct one. More precisely, we compute the following:

\begin{equation}
\begin{aligned}
s_1&=c^T \times M \times r\\
p_1&=sigmoid(s_1)
\end{aligned}
\end{equation}

where $M$ is a trainable weight matrix.

\subsubsection{Common Words Embedding} 

Another issue raised by~\citet{lowe2017training} is that direct word copying between the context and true response was not captured by Dual-LSTM model. In order to overcome this issue, we extract common word list from context and response, and feed common word embeddings to the LSTM network. We use the same word embedding layer as the other inputs of the Dual-LSTM model.

\begin{equation}
\begin{aligned}
& Common = LSTM(WE(context \cap response))\\
\end{aligned}
\end{equation}

Finally, we dense the hidden state of LSTM layer, and calculate the score of the corresponding response to be correct.

\begin{equation}
s_2=d^T \times Common,
\end{equation}

where $d$ is a trainable weight vector.

In the case of having both embedding layers, we concatenate the scores of common word embedding layer and Dual-LSTM layer and using sigmoid function to calculate the probability of the response to be the correct one:

\begin{equation}
p_2=sigmoid(\alpha_1 s_1+ \alpha_2 s_2)
\end{equation}

where $\alpha_1$ and $\alpha_2$ are trainable weights.

\subsection{Context-Response Common Words Frequency}
\label{crcwf}

This is a well-known fact that words which are more frequent (such as \emph{the}, \emph{is}, and \emph{that}) contain less information compared to more rare words (such as technical words in Ubuntu Corpus). We observe that when an utterance shares a rare word with some context, it is more probable that it is the correct response for the context. To capture this, we first create a table that stores word occurrence count. We use this table to calculate a variable that is the summation over reciprocal of common words' occurrences in a context and a response. More precisely; denoting by $n_{w_i}$ the number of times that word $w_i$ appeared in train dataset, for any context and response, we define $s(context,response)$ as follows:

\begin{equation}
\label{cucwfreq}
s(context,response)=\sum_{i: w_i \in cr} \frac{1}{n_{w_i}},
\end{equation}

where $cr=context \cap response$.

We note that unlike the TF-IDF model that computes TF-IDF vectors of both the context and the response and calculates the cosine similarity between the two vectors, our proposed layer only considers common words in the context and the response and is intended to reflect how informative a common word in context and response is.

\section{Experiment}
\label{experiment}

In this section, we provide details on our experiments including data preparation, experimental settings, the models we build using the networks and layers introduced in Section~\ref{method}, and training parameters and functions.

\subsection{Data Preparation}

We preprocessed the ubuntu dataset by normalizing every context and response using TweetMotif~\cite{o2010tweetmotif}. We use the tokenization function of TweetMotif to treat hashtags, @-replies, abbreviations, strings of punctuation, emoticons and unicode glyphs (e.g., musical notes) as tokens. In order to reproduce the original result, we kept all the train, validation, and test sets same as the original sets provided by~\citet{lowe2015ubuntu}.

\subsection{Experimental Setting}
\label{experimentalsetting}

The experiment was executed on Amazon AWS p2 xlarge machine with NVIDIA Tesla K80 GPU. We use Keras~\cite{chollet2015keras} to implement all our models. All models were trained using Root Mean Square Propagation optimizer (RMSProp)~\cite{hinton2012rmsprop} with learning rate set to $10^{-3}$ without decay. Also, the batch size is set to $256$ during training.

\subsection{Model Training}
\label{modeltraining}

\begin{figure}[h]
	\centering
    \includegraphics[width=\linewidth]{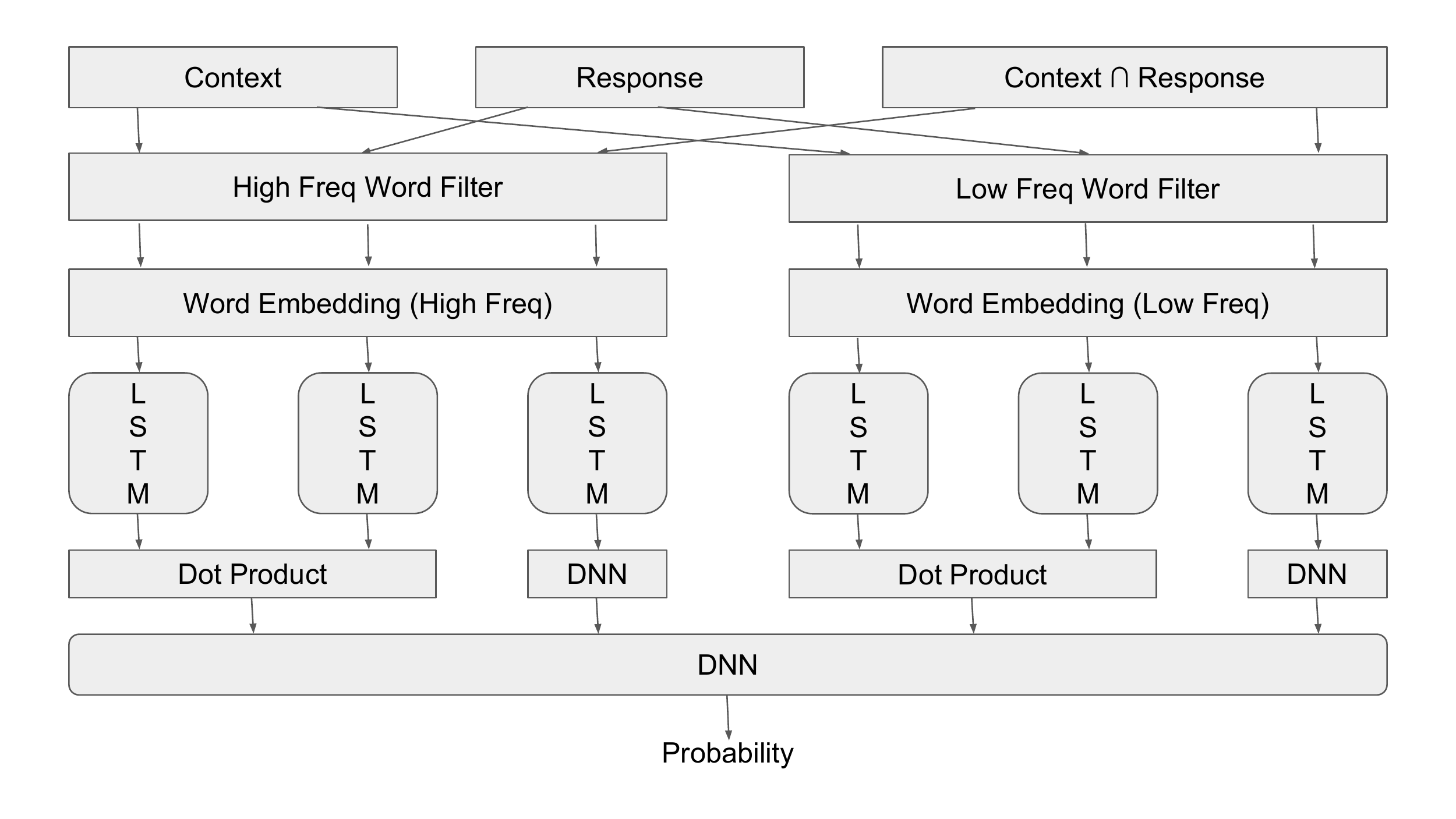}
    \caption{{\it Diagram of MFCW-LSTM model. $ Context \cap Response $ is the list of common words between context and response pair. Every single word in Context, Response, $ Context \cap Response $ is filtered by High Frequency Word Filter $(> 5)$ and Low Frequency Word Filter $(\leq 5)$, then feed to low frequency word embedding layer and high frequency word embedding layer, respectively. All the embedded results are fed into the correspond LSTM word-by-word. The LSTMs have tied weights for output of each word embedding layer respectively.}}
    \label{fig:MFCW-LSTM}
\end{figure}

We trained two different models with combination of few methods:

\begin{itemize}
\item Apply Multi Frequency Word Embedding and Common Word Embedding to LSTM structure. We refer to this model as MFCW-LSTM. Figure~\ref{fig:MFCW-LSTM} depicts the model.
\end{itemize}

\begin{figure}[h]
	\centering
    \includegraphics[width=\linewidth]{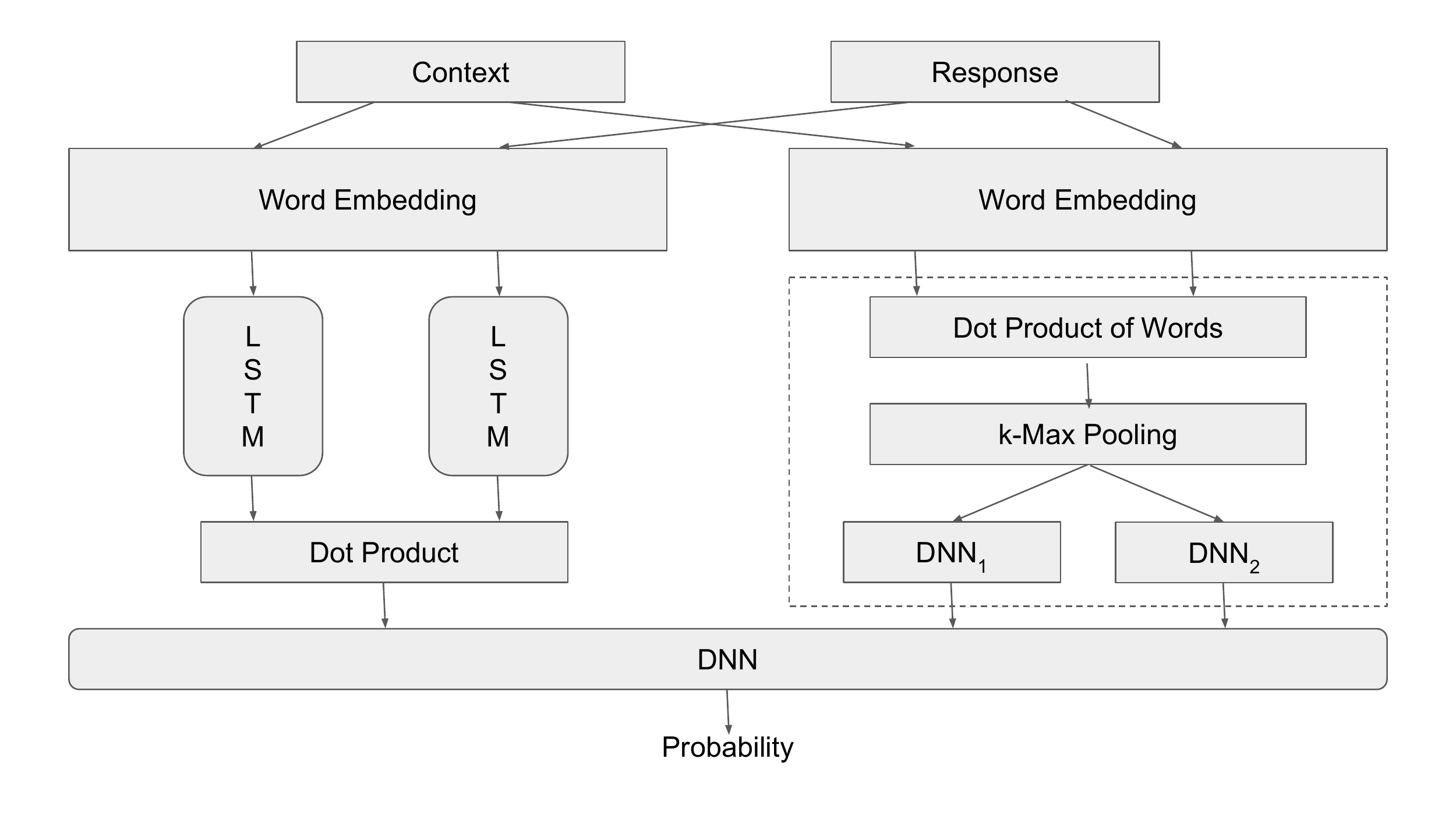}
    \caption{{\it Diagram of CCN-LSTM model. The LSTMs have tied weights, and the two word embedding layers do not share weights. The dashed box is the Context-Response Words Relation networks with Dot Product, k-Max Pooling, and Dense layers.}}
    \label{fig:CCN-LSTM}
\end{figure}

\begin{itemize}
\item Implement Cross Convolution Network and LSTM structure. High frequency layer of word embedding is the only embedding we use here. We refer to this model as CCN-LSTM. Figure~\ref{fig:CCN-LSTM} depicts the model.
\end{itemize}

All LSTM structures have $256$ hidden units. The maximum size of the context and response is set to $160$ words plus zero padding. The word embedding size is $300$ as in GloVe embedding.

For MFCW-LSTM, we tried different thresholds for high and low frequency word boundary, and found $5$ is the best threshold. We also learned parameters of LSTM with both shared weights and separate weights. For CCN-LSTM, we used two separated word embedding layers to feed words to the Context-Response Words Relation and LSTM networks. Parameter $k$ is set to $1$ and $2$ for different models; however, we did not see any improvement in models with $k=2$ over models with $k=1$.

To predict the label of a response for a context during training phase, we consider weighted sum of the response scores calculated by each of the networks in the considered model and apply sigmoid function that yields a number between $0$ and $1$. We then penalize the predicted label using square error loss function. The Context-Response Common Words Frequency layer is not used in the models during training. We discuss this issue more in the Result Section (Section~\ref{result}).

\subsection{Ensemble}
\label{ensemble}

Ensemble of multiple models can help us obtain better predictive performance than what we can get from any of the constituent models~\cite{opitz1999popular,polikar2006ensemble,sollich1996learning}. Similar to~\citet{kadlec2015improved}, we found that averaging the prediction result of multiple models gives a decent improvement. We found that the best classifier is ensemble of 16 MFCW-LSTMs and 4 CCN-LSTMs. 

\section{Result}
\label{result}

In this section we provide our experiments result in two subsections. Firstly, we report our models' performance and discuss the results. Afterwards, we provide comparison of our best model performance with previous work's.

\subsection{Evaluation}

We use the same evaluation metric as~\citet{lowe2015ubuntu}, namely $Recal@k$. Among $m=10$ response candidates provided in evaluation and test set, $1$ positive and $9$ negative responses are used. The model ranks $10$ responses, and prediction is considered correct if the correct response is in top $k$ candidates. we are reporting with $(m, k)$ of $(10, 1)$, $(10, 2)$ and $(10, 5)$.

\begin{figure}[h]
	\centering
    \includegraphics[width=\linewidth]{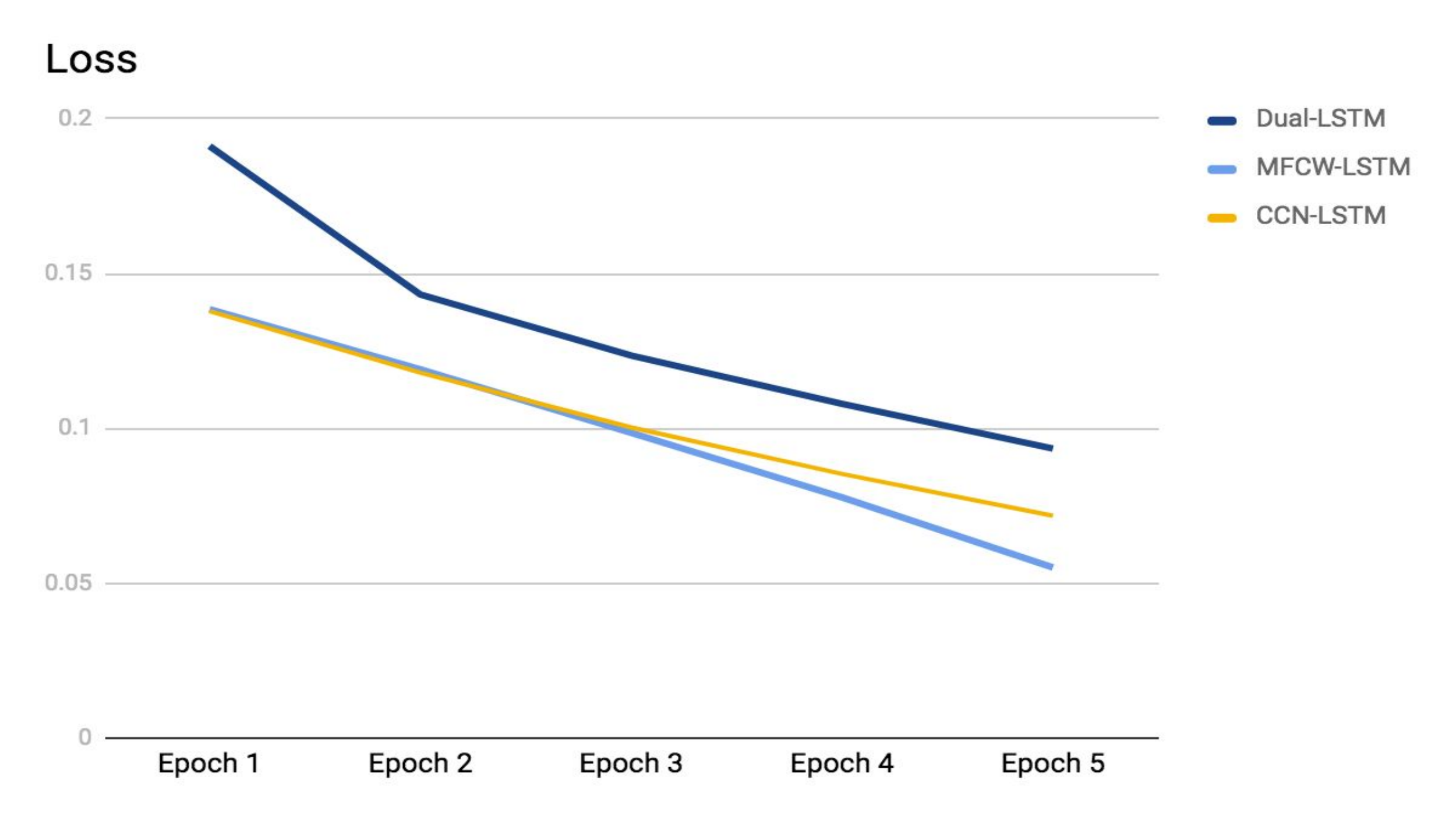}
    \caption{{\it Loss change for validation set.}}
    \label{fig:loss}
\end{figure}

\begin{figure}[h]
	\centering
    \includegraphics[width=\linewidth]{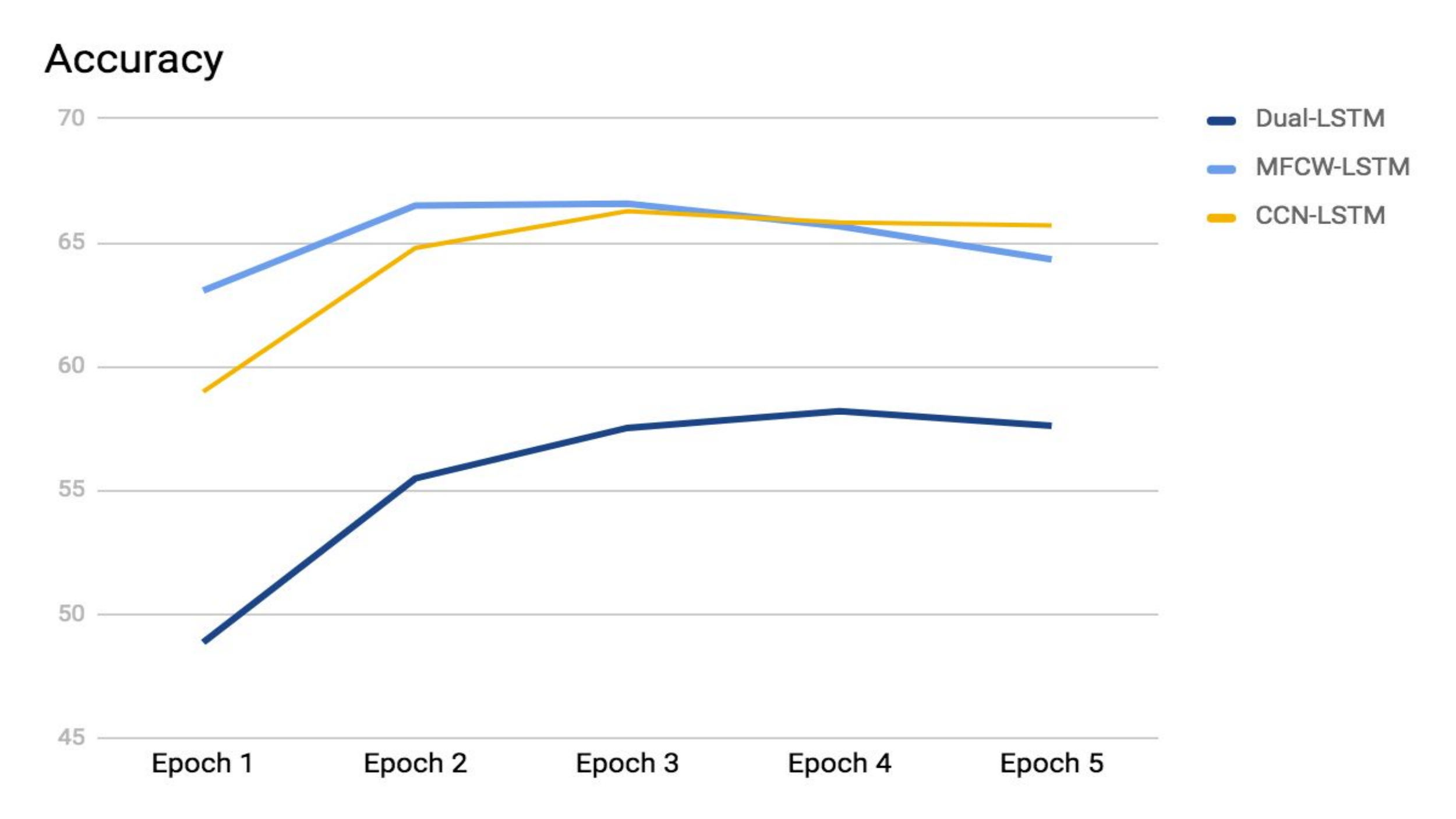}
    \caption{{\it Accuracy change for validation set. We found accuracy start decreasing after third epoch for various models.}}
    \label{fig:acc}
\end{figure}

We choose the best models using the accuracy on validation set (Figure~\ref{fig:loss} and Figure~\ref{fig:acc}). The performance of these models on test set are reported in Table~\ref{evaluation}. We reproduce the result of Dual-LSTM model as our baseline~\cite{lowe2015ubuntu}, and compare our models to this. We use same default hyper-parameter setting for Dual-LSTM as our other models, and initialize word embedding using Glove word vector~\cite{pennington2014glove}. The reproduced performance of Dual-LSTM model in here is better than original result~\cite{lowe2017training} since we are using high frequency word embedding and preprocessed dataset. 

MFCW-LSTM and CCN-LSTM are the models described in Subsection~\ref{modeltraining}. The CCN-LSTM model has $k=1$ and two dense parallel layers of linear and sigmoid in Context-Response Words Relation network as in Figure~\ref{fig:CCN-LSTM}. Both of these models outperform Dual-LSTM model by approximately 9\%. We also incorporate the MFCW-LSTM and CCN-LSTM to a single model, and the result is not better than any one of two.

We now investigate the effect of Context-Response Common Words Frequency (see Subsection~\ref{crcwf} for details.) in response prediction. The scaled scores computed using Equation~\ref{cucwfreq} are added to the resulting probabilities out of MFCW-LSTM and CCN-LSTM models in validation and test. Particularly, in order to predict the correct response on validation and test dataset, we first compute the score for each of the ten candidates responses using our model. After that we calculate $s(context,response)$ for each response, scale it, and add it to the model output. We then rank the responses based on this final score and choose the highest ranking as output. The scaling factor is optimized on the validation set. The combined models' results can be found in the MFCW-LSTM-CWF and CCN-LSTM-CWF rows (Table~\ref{evaluation}), where CWF stands for common words frequency. As we can see, CWF layer improve performance of both baseline models of MFCW-LSTM and CCN-LSTM. We see a noticeable improvement for CCN-LSTM which can be attributed to the fact that CCN-LSTM model does not include rare words in its decision.

\begin{table*}[t]
\caption{\label{evaluation} Result of our models on Ubuntu Dialogue Corpus test set for different recall measures. Numbers in bold is our best model performance.}
\vspace{1ex}
\centering
\begin{tabular}{|l|c|c|c|}\hline 
\textbf{Model} \rule{0pt}{1\normalbaselineskip} & \textbf{1 in 10R@1} & \textbf{1 in 10R@2} & \textbf{1 in 10R@5} \\ \hline
LSTM \rule{0pt}{1\normalbaselineskip} & 57.6\% & 75.3\% & 94.5\% \\
MFCW-LSTM & 66.5\% & 80.4\% & 95.4\% \\
CCN-LSTM & 66.3\% & 80.8\% & 95.6\% \\
MFCW-LSTM-CWF & 67.3\% & 81.2\% & 95.6\% \\ 
CCN-LSTM-CWF & 69.0\% & 82.2\% & 96.0\% \\
Ensemble \rule{1\normalbaselineskip}{0pt} & {\bf 72.7\%} & {\bf 85.8\%} & {\bf 97.1\%} \\
\hline
\end{tabular}
\end{table*}

\begin{table*}[t]
\caption{\label{comparison} Performance comparison of our best models and different recent papers. Numbers in bold mean that the improvement is statistically significant compared with previous baseline.}
\vspace{1ex}
\centering
\begin{tabular}{|l|c|c|c|}\hline
\textbf{Model} \rule{0pt}{1\normalbaselineskip} & \textbf{1 in10R@1} & \textbf{1 in10R@2} & \textbf{1 in10R@5} \\ \hline
Dual-LSTM \cite{lowe2017training} \rule{0pt}{1\normalbaselineskip}& 55.2\% & 72.1\% & 92.4\% \\
RNN-CNN \cite{baudivs2016sentence} & 67.2\% & 80.9\% & 95.6\% \\
r-LSTM \cite{xu2016incorporating} & 64.9\% & 78.5\% & 93.2\% \\ 
Ensemble \cite{kadlec2015improved} & 68.3\% &	81.8\% & 95.7\% \\
SMN \cite{wu2017sequential} & 72.6\% & 84.7\% & 96.2\% \\
Our Best Model & {\bf 72.7\%} & {\bf 85.8\%} & {\bf 97.1\%} \\
\hline
\end{tabular}
\end{table*}

\subsection{Comparison}
Table~\ref{comparison} shows performance comparison of our best models and different recent papers. Since we used the Ubuntu Dialogue Corpus v2 dataset, and we compare our result to other works based on the same version. Recently, \citet{wu2017sequential} achieved decent improvement over the previous state-of-the-art. As we can see our best model which is the ensemble model outperforms SMN~\cite{wu2017sequential} by $0.1\%$, $1.1\%$, and $0.9\%$ for 1 in10R@1, 1 in10R@2, and 1 in10R@5 metrics, respectively (Table ~\ref{comparison}). Therefore, we set a new state-of-the-art to 72.7\%, 85.8\% and 97.1\%.

\section{Conclusion and Future Work}
\label{conclusionfuture}

In this paper, we considered the problem of next response selection for multi-turn conversation. Motivating by the large gap between machine and expert performances on this task for Ubuntu Dialogue Corpus, we presented new networks and layers and we evaluated our models using this dataset. We proposed Cross Convolution Network (CCN) that is potentially useful for the general task of classifying a pair of objects. We implemented CCN combined with LSTM as one of our single model. The CNN tries to capture word level information on word pairs of the context and response, while the LSTM captures the information on the entire context and the entire response. We also investigated the effect of Multi Frequency Word Embedding and Common Words Embedding combined with LSTM as our other model. The Multi Frequency Word Embedding tries to embed both rare and frequent words in an efficient way, and it is able to capture important low frequency key words without increasing too much computational complexity. Our experimental results showed a promising improvement over previous models; specifically when we ensemble our models to select the next response. 

For future work, we will explore the fusion of these findings in other multi-turn response selection dataset and other related problems, and evaluate whether the gains achieved here are orthogonal to other methods for improving performance. We also see the potential of extending our framework to generative models for dialogue systems.

\section{Acknowledgments}

We thank Morten Pedersen and David Guy Brizan for their contributions to this study. We gratefully acknowledge financial support for this work by AOL of OATH. 

\bibliography{naaclhlt2018}

\begin{thebibliography}{}
\expandafter\ifx\csname natexlab\endcsname\relax\def\natexlab#1{#1}\fi

\bibitem[{Baudi{\v{s}} et~al.(2016)Baudi{\v{s}}, Pichl, Vysko{\v{c}}il, and
  {\v{S}}ediv{\`y}}]{baudivs2016sentence}
Petr Baudi{\v{s}}, Jan Pichl, Tom{\'a}{\v{s}} Vysko{\v{c}}il, and Jan
  {\v{S}}ediv{\`y}. 2016.
\newblock Sentence pair scoring: Towards unified framework for text
  comprehension.
\newblock {\em arXiv preprint arXiv:1603.06127\/} .

\bibitem[{Cho et~al.(2014)Cho, Van~Merri{\"e}nboer, Gulcehre, Bahdanau,
  Bougares, Schwenk, and Bengio}]{cho2014learning}
Kyunghyun Cho, Bart Van~Merri{\"e}nboer, Caglar Gulcehre, Dzmitry Bahdanau,
  Fethi Bougares, Holger Schwenk, and Yoshua Bengio. 2014.
\newblock Learning phrase representations using rnn encoder-decoder for
  statistical machine translation.
\newblock {\em arXiv preprint arXiv:1406.1078\/} .

\bibitem[{Chollet et~al.(2015)}]{chollet2015keras}
Fran\c{c}ois Chollet et~al. 2015.
\newblock Keras.
\newblock \url{https://github.com/fchollet/keras}.

\bibitem[{Graves et~al.(2013)Graves, Mohamed, and Hinton}]{graves2013speech}
Alex Graves, Abdel-rahman Mohamed, and Geoffrey Hinton. 2013.
\newblock Speech recognition with deep recurrent neural networks.
\newblock In {\em Acoustics, speech and signal processing (icassp), 2013 ieee
  international conference on\/}. IEEE, pages 6645--6649.

\bibitem[{Henderson et~al.(2013)Henderson, Thomson, and
  Young}]{henderson2013deep}
Matthew Henderson, Blaise Thomson, and Steve~J Young. 2013.
\newblock Deep neural network approach for the dialog state tracking challenge.
\newblock In {\em SIGDIAL Conference\/}. pages 467--471.

\bibitem[{Hinton et~al.(2012)Hinton, Srivastava, and
  Swersky}]{hinton2012rmsprop}
G~Hinton, N~Srivastava, and K~Swersky. 2012.
\newblock Rmsprop: Divide the gradient by a running average of its recent
  magnitude.
\newblock {\em Neural networks for machine learning, Coursera lecture 6e\/} .

\bibitem[{Hochreiter and Schmidhuber(1997)}]{hochreiter1997long}
Sepp Hochreiter and J{\"u}rgen Schmidhuber. 1997.
\newblock Long short-term memory.
\newblock {\em Neural computation\/} 9(8):1735--1780.

\bibitem[{Ji et~al.(2014)Ji, Lu, and Li}]{ji2014information}
Zongcheng Ji, Zhengdong Lu, and Hang Li. 2014.
\newblock An information retrieval approach to short text conversation.
\newblock {\em arXiv preprint arXiv:1408.6988\/} .

\bibitem[{Kadlec et~al.(2015)Kadlec, Schmid, and
  Kleindienst}]{kadlec2015improved}
Rudolf Kadlec, Martin Schmid, and Jan Kleindienst. 2015.
\newblock Improved deep learning baselines for ubuntu corpus dialogs.
\newblock {\em arXiv preprint arXiv:1510.03753\/} .

\bibitem[{Kalchbrenner et~al.(2014)Kalchbrenner, Grefenstette, and
  Blunsom}]{kalchbrenner2014convolutional}
Nal Kalchbrenner, Edward Grefenstette, and Phil Blunsom. 2014.
\newblock A convolutional neural network for modelling sentences.
\newblock {\em arXiv preprint arXiv:1404.2188\/} .

\bibitem[{Lowe et~al.(2015)Lowe, Pow, Serban, and Pineau}]{lowe2015ubuntu}
Ryan Lowe, Nissan Pow, Iulian Serban, and Joelle Pineau. 2015.
\newblock The ubuntu dialogue corpus: A large dataset for research in
  unstructured multi-turn dialogue systems.
\newblock {\em arXiv preprint arXiv:1506.08909\/} .

\bibitem[{Lowe et~al.(2016)Lowe, Serban, Noseworthy, Charlin, and
  Pineau}]{lowe2016evaluation}
Ryan Lowe, Iulian~V Serban, Mike Noseworthy, Laurent Charlin, and Joelle
  Pineau. 2016.
\newblock On the evaluation of dialogue systems with next utterance
  classification.
\newblock {\em arXiv preprint arXiv:1605.05414\/} .

\bibitem[{Lowe et~al.(2017)Lowe, Pow, Serban, Charlin, Liu, and
  Pineau}]{lowe2017training}
Ryan~Thomas Lowe, Nissan Pow, Iulian~Vlad Serban, Laurent Charlin, Chia-Wei
  Liu, and Joelle Pineau. 2017.
\newblock Training end-to-end dialogue systems with the ubuntu dialogue corpus.
\newblock {\em Dialogue \& Discourse\/} 8(1):31--65.

\bibitem[{Lu and Li(2013)}]{lu2013deep}
Zhengdong Lu and Hang Li. 2013.
\newblock A deep architecture for matching short texts.
\newblock In {\em Advances in Neural Information Processing Systems\/}. pages
  1367--1375.

\bibitem[{Mikolov et~al.(2013)Mikolov, Sutskever, Chen, Corrado, and
  Dean}]{mikolov2013distributed}
Tomas Mikolov, Ilya Sutskever, Kai Chen, Greg~S Corrado, and Jeff Dean. 2013.
\newblock Distributed representations of words and phrases and their
  compositionality.
\newblock In {\em Advances in neural information processing systems\/}. pages
  3111--3119.

\bibitem[{O'Connor et~al.(2010)O'Connor, Krieger, and Ahn}]{o2010tweetmotif}
Brendan O'Connor, Michel Krieger, and David Ahn. 2010.
\newblock Tweetmotif: Exploratory search and topic summarization for twitter.
\newblock In {\em ICWSM\/}. pages 384--385.

\bibitem[{Opitz and Maclin(1999)}]{opitz1999popular}
David~W Opitz and Richard Maclin. 1999.
\newblock Popular ensemble methods: An empirical study.
\newblock {\em J. Artif. Intell. Res.(JAIR)\/} 11:169--198.

\bibitem[{Pennington et~al.(2014)Pennington, Socher, and
  Manning}]{pennington2014glove}
Jeffrey Pennington, Richard Socher, and Christopher~D Manning. 2014.
\newblock Glove: Global vectors for word representation.
\newblock In {\em EMNLP\/}. volume~14, pages 1532--1543.

\bibitem[{Polikar(2006)}]{polikar2006ensemble}
Robi Polikar. 2006.
\newblock Ensemble based systems in decision making.
\newblock {\em IEEE Circuits and systems magazine\/} 6(3):21--45.

\bibitem[{Shang et~al.(2015)Shang, Lu, and Li}]{shang2015neural}
Lifeng Shang, Zhengdong Lu, and Hang Li. 2015.
\newblock Neural responding machine for short-text conversation.
\newblock {\em arXiv preprint arXiv:1503.02364\/} .

\bibitem[{Sollich and Krogh(1996)}]{sollich1996learning}
Peter Sollich and Anders Krogh. 1996.
\newblock Learning with ensembles: How overfitting can be useful.
\newblock In {\em Advances in neural information processing systems\/}. pages
  190--196.

\bibitem[{Sordoni et~al.(2015)Sordoni, Galley, Auli, Brockett, Ji, Mitchell,
  Nie, Gao, and Dolan}]{sordoni2015neural}
Alessandro Sordoni, Michel Galley, Michael Auli, Chris Brockett, Yangfeng Ji,
  Margaret Mitchell, Jian-Yun Nie, Jianfeng Gao, and Bill Dolan. 2015.
\newblock A neural network approach to context-sensitive generation of
  conversational responses.
\newblock {\em arXiv preprint arXiv:1506.06714\/} .

\bibitem[{Sutskever et~al.(2014)Sutskever, Vinyals, and
  Le}]{sutskever2014sequence}
Ilya Sutskever, Oriol Vinyals, and Quoc~V Le. 2014.
\newblock Sequence to sequence learning with neural networks.
\newblock In {\em Advances in neural information processing systems\/}. pages
  3104--3112.

\bibitem[{Wan et~al.(2016)Wan, Lan, Guo, Xu, Pang, and Cheng}]{wan2016deep}
Shengxian Wan, Yanyan Lan, Jiafeng Guo, Jun Xu, Liang Pang, and Xueqi Cheng.
  2016.
\newblock A deep architecture for semantic matching with multiple positional
  sentence representations.
\newblock In {\em AAAI\/}. pages 2835--2841.

\bibitem[{Wang et~al.(2015)Wang, Lu, Li, and Liu}]{wang2015syntax}
Mingxuan Wang, Zhengdong Lu, Hang Li, and Qun Liu. 2015.
\newblock Syntax-based deep matching of short texts.
\newblock {\em arXiv preprint arXiv:1503.02427\/} .

\bibitem[{Wen et~al.(2015{\natexlab{a}})Wen, Gasic, Kim, Mrksic, Su, Vandyke,
  and Young}]{wen2015stochastic}
Tsung-Hsien Wen, Milica Gasic, Dongho Kim, Nikola Mrksic, Pei-Hao Su, David
  Vandyke, and Steve Young. 2015{\natexlab{a}}.
\newblock Stochastic language generation in dialogue using recurrent neural
  networks with convolutional sentence reranking.
\newblock {\em arXiv preprint arXiv:1508.01755\/} .

\bibitem[{Wen et~al.(2015{\natexlab{b}})Wen, Gasic, Mrksic, Su, Vandyke, and
  Young}]{wen2015semantically}
Tsung-Hsien Wen, Milica Gasic, Nikola Mrksic, Pei-Hao Su, David Vandyke, and
  Steve Young. 2015{\natexlab{b}}.
\newblock Semantically conditioned lstm-based natural language generation for
  spoken dialogue systems.
\newblock {\em arXiv preprint arXiv:1508.01745\/} .

\bibitem[{Wu et~al.(2017)Wu, Wu, Xing, Zhou, and Li}]{wu2017sequential}
Yu~Wu, Wei Wu, Chen Xing, Ming Zhou, and Zhoujun Li. 2017.
\newblock Sequential matching network: A new architecture for multi-turn
  response selection in retrieval-based chatbots.
\newblock In {\em Proceedings of the 55th Annual Meeting of the Association for
  Computational Linguistics (Volume 1: Long Papers)\/}. volume~1, pages
  496--505.

\bibitem[{Xian and Tian(2017)}]{xian2017self}
Yang Xian and Yingli Tian. 2017.
\newblock Self-guiding multimodal lstm-when we do not have a perfect training
  dataset for image captioning.
\newblock {\em arXiv preprint arXiv:1709.05038\/} .

\bibitem[{Xu et~al.(2016)Xu, Liu, Wang, Sun, and Wang}]{xu2016incorporating}
Zhen Xu, Bingquan Liu, Baoxun Wang, Chengjie Sun, and Xiaolong Wang. 2016.
\newblock Incorporating loose-structured knowledge into lstm with recall gate
  for conversation modeling.
\newblock {\em arXiv preprint arXiv:1605.05110\/} .

\bibitem[{Yan et~al.(2016)Yan, Song, and Wu}]{yan2016learning}
Rui Yan, Yiping Song, and Hua Wu. 2016.
\newblock Learning to respond with deep neural networks for retrieval-based
  human-computer conversation system.
\newblock In {\em Proceedings of the 39th International ACM SIGIR conference on
  Research and Development in Information Retrieval\/}. ACM, pages 55--64.

\bibitem[{Yih et~al.(2015)Yih, Chang, He, and Gao}]{yih2015semantic}
Scott Wen-tau Yih, Ming-Wei Chang, Xiaodong He, and Jianfeng Gao. 2015.
\newblock Semantic parsing via staged query graph generation: Question
  answering with knowledge base .

\bibitem[{Yu et~al.(2014)Yu, Hermann, Blunsom, and Pulman}]{yu2014deep}
Lei Yu, Karl~Moritz Hermann, Phil Blunsom, and Stephen Pulman. 2014.
\newblock Deep learning for answer sentence selection.
\newblock {\em arXiv preprint arXiv:1412.1632\/} .

\bibitem[{Zhou et~al.(2016)Zhou, Dong, Wu, Zhao, Yu, Tian, Liu, and
  Yan}]{zhou2016multi}
Xiangyang Zhou, Daxiang Dong, Hua Wu, Shiqi Zhao, Dianhai Yu, Hao Tian, Xuan
  Liu, and Rui Yan. 2016.
\newblock Multi-view response selection for human-computer conversation.
\newblock In {\em EMNLP\/}. pages 372--381.

\end{thebibliography}
\bibliographystyle{acl_natbib}

\end{document}